\documentclass{article}

\usepackage[final,nonatbib]{neurips_2022}

\usepackage[utf8]{inputenc} 
\usepackage[T1]{fontenc}    
\usepackage{hyperref}       
\usepackage{url}            
\usepackage{booktabs}       
\usepackage{amsfonts}       
\usepackage{makecell}
\usepackage{nicefrac}       
\usepackage{microtype}      
\usepackage{xcolor}         
\usepackage{graphicx}
\usepackage{multirow}
\title{Multi-stream Cell Segmentation with Low-level Cues for Multi-modality Images}

\author{%
  Wei Lou$^{*}$, Xinyi Yu\thanks{Equal Contributions}, Chenyu Liu , Xiang Wan, Guanbin Li, Siqi Liu, Haofeng Li\thanks{Corresponding Author} \\
  Shenzhen Research Institute of Big Data,\\
  The Chinese University of Hong Kong (Shenzhen)\\
  \texttt{lhaof@sribd.cn}\\
}

\begin{document}

\maketitle

\begin{abstract}
Cell segmentation for multi-modal microscopy images remains a challenge due to the complex textures, patterns, and cell shapes in these images. To tackle the problem, we first develop an automatic cell classification pipeline to label the microscopy images based on their low-level image characteristics, and then train a classification model based on the category labels. Afterward, we train a separate segmentation model for each category using the images in the corresponding category. Besides, we further deploy two types of segmentation models to segment cells with roundish and irregular shapes respectively. Moreover, an efficient and powerful backbone model is utilized to enhance the efficiency of our segmentation model. Evaluated on the Tuning Set of NeurIPS 2022 Cell Segmentation Challenge, our method achieves an F1-score of 0.8795 and the running time for all cases is within the time tolerance.
\end{abstract}

\section{Introduction}
\label{intro}
Cell segmentation is a common task in digital biomedical analysis. The goal of cell segmentation is to label the contour of each cell. Segmenting cells is helpful for many healthcare and life science applications, such as tumor detection, medicine customization, and the understanding of cell population heterogeneity ~\cite{prangemeier2020attention,lee2022cellseg,sun2020deep,prangemeier2020multiclass,leygeber2019analyzing,lou2022pixel,zhou2021ssmd}. The common challenge for cell segmentation in practice is that the microscopy image datasets are usually comprised of multi-modal images. The images from different modalities could have various textures, patterns, cell size/shape, and usually lack the annotations of modality. It is difficult to achieve satisfying generalization on a multi-modal cell segmentation dataset using only a single model.

Therefore, we propose that the images with similar low-level features are suitable to form a group. It is conducive to the convergence by training a segmentation model with such an image group. To achieve robust image classification for testing images, we automatically label all the training images into four groups (namely, categories) in an unsupervised manner, and then train a simple image classification model. Each testing image is sent into the classifier to predict the image category, and then sent into the cell segmentation model that is trained for that category. Besides, we find that different segmentation methods show the performance diversity on the images with various low-level features. The methods regressing the distances to polygon boundaries may be unsuitable for segmenting the cells with irregular shape and low convexity. Therefore, we design a class-wise multi-stream cell segmentation framework to tackle the difficulties of the task. In the proposed framework, for most of the classes with roundish cells, we deploy Stardist~\cite{schmidt2018cell} models to segment the cell instances. For the cell class with irregular and high-concavity shapes, we use a  Hover-net~\cite{graham2019hover} model. All these models are first trained using the whole training set, and then fine-tuned with the images of the corresponding class that is assigned to the model. Moreover, we deploy a simple and efficient backbone model ConvNeXt~\cite{liu2022convnet} to raise the performance and efficiency of the segmentation model. In this paper, our contributions can be summarized as follows:
\begin{itemize}
\item We design an automatic cell image classification pipeline to classify the multi-modal datasets into four categories based on low-level visual cues, and then train a robust CNN classifier.
\item We integrate two advanced segmentation models for different image categories, by utilizing their advantages in segmenting cells of low- or high-convexity shapes.
\item We deploy an efficient backbone model that boosts the segmentation performance and reduces the inference time.
\end{itemize}

\section{Method}
\label{method}
The proposed method contains two parts: 1). Automatic cell image classification; 2). Class-wise multi-stream cell segmentation. In the step of automatic cell classification, all labeled or unlabelled images are classified into four categories based on low-level patterns such as color and cell size. In the step of class-wise multi-stream cell segmentation, we use a specific model for the cells of each category. In the training stage, all the models are trained with the whole dataset, and fine-tuned with the images of the corresponding class, respectively.

\subsection{Automatic cell image classification}
\label{sec.2.2}
The whole automatic cell image classification pipeline can be described as three steps: 1). Pseudo segmentation label synthesis. 2). Unsupervised cell image classification. 3). Deep classification model training. In this work, we utilize four datasets, the dataset of the Cell Segmentation Challenge and three public datasets (Cellpose~\cite{stringer2021cellpose}, Omnipose~\cite{cutler2022omnipose}, Sartorius~\cite{Sartorius}).

\begin{figure*}[h]
\centering
\includegraphics[width=0.85\linewidth]{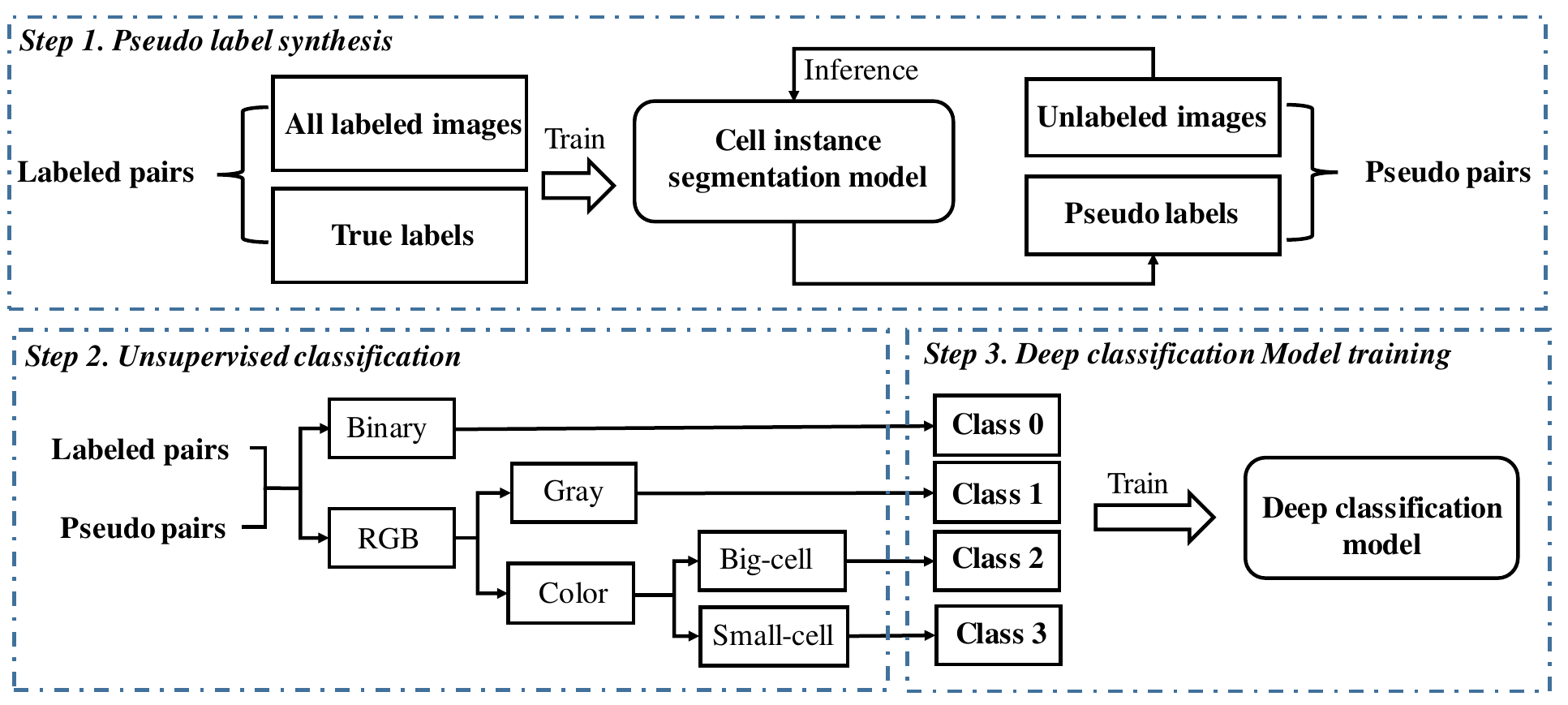}
\caption{The automatic cell image classification.}
\label{fig_1}
\end{figure*}

The unsupervised classification for cell images is based on low-level features including the characteristics of cell masks. To make use of unlabeled images, we need to synthesize pseudo labels~\cite{lee2013pseudo,lou2022pixel} of cell instance segmentation, which is shown as Step 1 in Figure \ref{fig_1}. In the pseudo-labeling method, a fully-supervised segmentation model is trained with all the labeled data, including three public datasets. Then the trained segmentation model can predict cell instances for each unlabeled image. Those predicted cell instances are pseudo labels. In the step of pseudo label synthesis, we adopt a modified Stardist model called ConvNeXt-Stardist as the supervised segmentation model to produce the pseudo labels. The details of the segmentation model are in Sec.~\ref{sec.2.3.2}.

In the step of unsupervised classification, we first divide the labeled and pseudo pairs from all the datasets into binary and RGB images, according to the number of image channels. All the binary images with a single channel are classified into Class 0. Second, we convert these RGB images to HSV color space, and separate them into gray and color images using S (saturation) and V (value). If the mean saturation of an image is larger than $\theta$ and the mean value of the image is within a range of ($\alpha_s$,$\alpha_l$), then it is classified as a gray image (Class 1). Otherwise, it is regarded as a color image. Third, we categorize the color images according to the maximum cell area in each image. If the maximum cell area of a color image is larger than $\sigma$ pixels, the color image is classified into the category of large-cell images (Class 2). Otherwise, the color image is categorized into the group of small-cell images (Class 3). Finally, all the images in the four datasets are separated into binary images (Class0), gray images (Class1), large-cell images (Class2), and small-cell images (Class3) automatically.

In the step of deep classification model training, we train a ResNet18~\cite{he2016deep} image classifier by using the unsupervised classification results as labels so that the CNN classifier can robustly predict the category of testing images. We set the output channels of ResNet18 to 4, and take the modified ResNet18 as the image classifier. The input of the classification model is an image of shape $3\times 224\times 224$ and the output contains the probabilities of belonging to each category. We divide all the images labeled in the unsupervised classification stage into training and validation sets to train the deep CNN classifier. 

\subsection{Class-wise multi-stream cell segmentation framework}

For microscopy images of different categories, we design a class-wise multi-stream cell segmentation framework to better utilize the advantages of different segmentation models. In the proposed framework, we modify two instance segmentation methods \cite{schmidt2018cell,graham2019hover} and integrate them with a powerful but efficient backbone network ConvNeXt~\cite{liu2022convnet}. The modified models are called ConvNeXt-Stardist and ConvNeXt-Hover, respectively.
\begin{figure*}[h]
\centering
\includegraphics[width=0.8\linewidth]{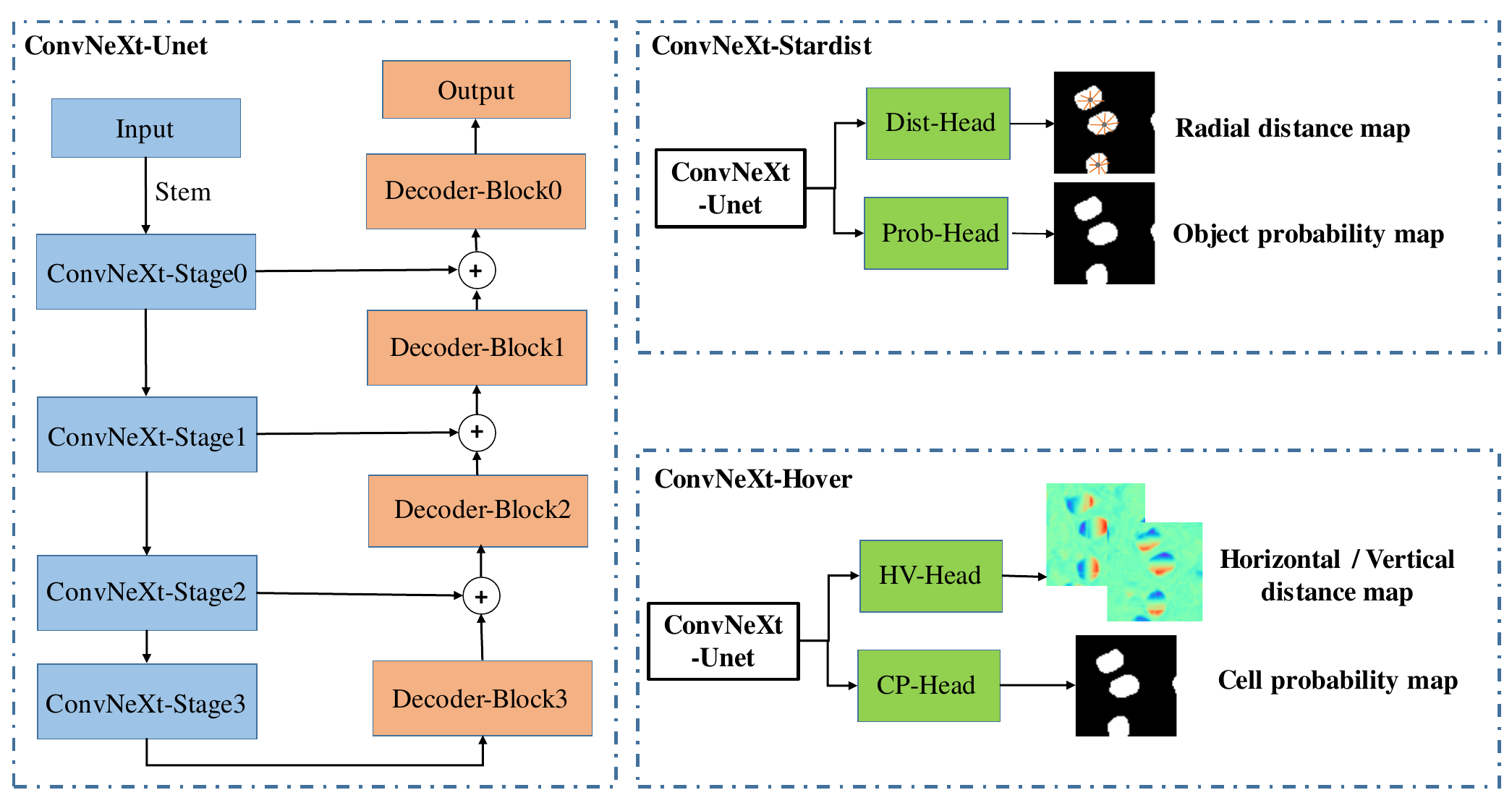}
\caption{The architectures of ConvNeXt-Stardist and ConvNeXt-Hover. Note that the ConvNeXt-Unets used in these two models are independent and have different model weights.}
\label{fig_2}
\end{figure*}

\textbf{ConvNeXt-Unet}
We deploy a U-net~\cite{ronneberger2015u,zhou2021ssmd} architecture as the segmentation framework, as shown in Figure~\ref{fig_2}. To maintain the segmentation performance and efficiency, we utilize the ConvNeXt-small~\cite{liu2022convnet} network as the encoder. The encoder consists of a stem layer and four ConvNeXt stages. The decoder contains four Conv-Relu-BN blocks. The number of ConvNeXt blocks in four stages is [3, 3, 27, 3]. The stem layer is a $4 \times 4$ convolution layer with a stride of 4. Each Conv-ReLU-BN block contains two $3 \times 3$ convolution layers. Each convolution layer is followed by a ReLU activation function and a batch normalization layer. The input of the ConvNeXt-Unet is of shape $3 \times H \times W$ and the output feature is of shape $C \times \frac{H}{4} \times \frac{W}{4}$.

\textbf{ConvNeXt-Stardist}\label{sec.2.3.2}
We observe that most of the cells in Class 0,2,3 are of roundish shapes that can be well described or approximated by a star-convex polygon. Thus, we combine the prediction heads of the Stardist~\cite{schmidt2018cell} model with the ConvNeXt-Unet feature extractor to segment cell instances. The ConvNeXt-Stardist model utilizes two prediction heads (Prob head and Dist head) to output an object probability map $p$ of shape $1 \times H \times W$ and a distance map $d$ of shape $R \times H \times W$. $R$ is the number of rays to build a star-convex polygon. $H$ and $W$ denote the height and width of the input image. The object probability of a pixel is defined as the normalized smallest Euclidean distance to the background region. $d_{i} \in d$ is a vector of shape $R \times 1$, which contains $R$ predicted Euclidean distances between the $i^{th}$ pixel and the object boundary in $R$ directions. Each prediction head contains an $\times 4$ upsampling layer, a $1 \times 1$ convolution layer, and a nonlinear activation function. The Prob head and Dist head adopt Sigmoid and ReLU as the activation function, respectively. 

The loss function consists of two parts. The first part is the cross entropy (CE) loss: $L_{CE}=-\frac{1}{n} \sum_{i=1}^N P_{i} \log P^*_{i}$, it is computed between the predicted probability map $P$ and the ground truth probability $P^{*}$. The second part consists of the Dice loss and MAE loss: $L_{Dice} =1-\frac{2 \times \sum_{i=1}^N\left(d^*_i \times d_i\right)+\epsilon}{\sum_{i=1}^N d^*_i+\sum_{i=1}^N d_i+\epsilon}$ and $L_{MAE}=\frac{1}{N}\sum_{i=1}^n\left|d^*_i-d_i\right|$. $N$ denotes the number of pixels in the input image. They are calculated between the predicted distance map $d$ and the ground truth distance map $d^{*}$. The weights for CE loss, Dice loss and MAE loss are 1, 1, and 0.3, respectively. For post-processing, we perform non-maximum suppression (NMS) \cite{ren2015faster} to preserve the cell objects with high object probabilities and remove the highly overlapping cells.

\textbf{ConvNeXt-Hover}
Since Stardist is designed for the cells with convex-polygon shapes, it is unpromising to segment irregular cells that account for a considerable proportion of Class 1. Thus, for Class 1 we resort to HoverNet \cite{graham2019hover} that segments a cell via its horizontal and vertical distance maps (HV map). These two distance maps measure the horizontal and vertical distances from each cell pixel to its cell center, respectively. Both high and low-convexity cell shapes could be predicted by HoverNet. To exploit the strength of the ConvNeXt-small backbone, we replace the Stardist heads in ConvNeXt-Stardist with the HoverNet heads to build the ConvNeXt-Hover network, instead of using multiple decoders as the original HoverNet. The prediction heads of ConvNeXt-Hover produce a cell pixel (CP) map of shape $2\times H\times W$, and the distance maps of shape $2\times H\times W$. The CP map predicts the probability of each pixel being within a cell. Each prediction head in ConvNeXt-Hover contains a $\times 4$ upsampling layer and a $1\times 1$ convolution layer. The prediction heads of ConvNeXt-Hover are the HV head and CP head, which adopt Identity and Softmax as the activation function, respectively.

The same CE loss and Dice loss used in ConvNeXt-Hover are computed for the predicted CP map. The mean squared error (MSE) loss: $\mathcal{L}_{MSE}=\frac{1}{N} \sum_{i=1}^N\left(p^{hv}_i-\Gamma_i\right)^2$  is calculated for the predicted horizontal/vertical distance maps $p^{hv}$ and the ground truth $\Gamma$. $N$ denotes the number of pixels in the input image. The mean squared gradient error (MSGE) loss is defined as $\mathcal{L}_{MSGE} =\frac{1}{m} \sum_{i \in M}\left(\nabla_x\left(p^{hv}_{i, x}\right)-\nabla_x\left(\Gamma_{i, x}\right)\right)^2 +\frac{1}{m} \sum_{i \in M}\left(\nabla_y\left(p^{hv}_{i, y}\right)-\nabla_y\left(\Gamma_{i, y}\right)\right)^2$. $m$ denotes the number of nuclei pixels and $M$ denotes the set containing all these nuclei pixels in the input image. The MSGE loss is computed between the gradients of predicted HV maps and the ground truth, $\nabla_x$ and $\nabla_y$ are the gradients in the horizontal and vertical directions, respectively. The weights for the above four losses are set to 1. For post-processing, we perform the marker-controlled watershed algorithm~\cite{cheng2008segmentation} using the predicted CP map and HV maps to separate overlapping cells.

\subsection{Training and inference scheme}
We train four cell segmentation models to segment the four categories, respectively. For Class 0, 2, 3, we first pre-train a ConvNeXt-Stardist model using all the labeled images. Then we utilize the pre-trained model weights to fine-tune three ConvNeXt-Stardist networks (denoted as Model 0, 2, 3) using the images of the corresponding category (Class 0, 2, 3), respectively. For Class 1, we pre-train a ConvNeXt-Hover model with all labeled images, and then fine-tune the model (denoted as Model 1) using only the images of Class 1. During the inference stage, a testing image is classified by the deep classification model trained in Sec.~\ref{sec.2.2}, and then is processed by the trained segmentation model that has the corresponding categorical label with the testing image.

\section{Experiments}
\label{headings}

\subsection{Dataset}
\begin{table}[!htbp]
\centering
\caption{The datasets and the numbers of samples used in the competition.}
\begin{tabular}{c|ccc}
\label{table:dataset}
{Datasets} & {Training} & {Validation} & {Tuning} \\
\hline
Cell-seg competition& 800  & 200          & 101   \\
\hline
Cellpose & 608  & -  & -  \\
\hline
Omnipose  & 735  & -  & -   \\
\hline
Sartorius  & 606 & - & - \\
\hline
\end{tabular}
\end{table}

During the training, We combine the dataset provided by the challenge organizer with 3 different public datasets, Cellpose~\cite{stringer2021cellpose}, Omnipose~\cite{cutler2022omnipose} and Sartorius
\footnote[1]{https://www.kaggle.com/competitions/sartorius-cell-instance-segmentation/overview}. The dataset provided by the organizer consists of four microscopy modalities including Brightfield (300 images), Fluorescent (300 images), Phase-contrast (200 images) and Differential interference contrast (200 images). The Cellpose dataset~\cite{stringer2021cellpose} consists of fluorescently labeled proteins that localized to the cytoplasm with or without 4,6-diamidino-2-phenylindole- (DAPI-) stained nuclei in a separate channel (316 images), 50 images of cells from brightfield microscopy, 58 images of membrane-labeled cells and 86 microscopy images of other types. Besides, Cellpose includes 98 nonmicroscopy images such as fruits, rocks and jellyfish for better generalization ability. The Omnipose dataset~\cite{cutler2022omnipose} is based on two publicly available microscopy datasets~\cite{javer2018open,ljosa2012annotated}. The Satorius dataset is from a Kaggle competition and consists of three neuronal cell types, including 320 Cortical neurons (Cort), 155 Shsy5y, and 131 astrocytes (Astro) samples.

We randomly select 200 samples from the dataset provided by the competition organizer as the validation set, and the rest samples plus the public dataset are used as the training set. The tuning set provided by the competition is used as a testing set. Resulting in 2749 training cases, 200 validation cases, and 101 testing cases. In our implementations, we adopt the Imagenet-pretrained~\cite{deng2009imagenet} model weights to initialize the ConvNeXt encoder.

\subsection{Evaluation Metrics}
\textbf{Mean F1 score} 
The metric is defined as the harmonic mean of the model’s \textit{precision} and \textit{recall}. \textit{Precision} is the fraction of true positive samples among the predicted cell instances. \textit{Recall} is the fraction of true positive samples among the ground truth cells. A true positive sample is a predicted cell instance whose largest IOU with some ground truth cell is more than 0.5.

\textbf{Real running time} The running time of starting a docker container and inferring an image.

\textbf{Out-of-tolerance running time} If the real running time is within the time tolerance, then the out-of-tolerance running time is 0. Otherwise, the out-of-tolerance (Out-of-Tol) running time is obtained by subtracting the time tolerance from the real running time. The time tolerance is proportional to the image size if the image size is more than 1,000,000 otherwise 10 seconds.

\subsection{Implementation details}
The development environments and requirements are shown in Table~\ref{table:env}. Some training details are shown in Table~\ref{table:training}.

\begin{table}[!htbp]
\centering
\small 
\caption{Development environments and requirements.}\label{table:env}
\begin{tabular}{ll}
\hline
System       & Linux\\
\hline
CPU   & Intel(R) Xeon(R) Gold 6248 CPU @ 2.50GHz \\
\hline
RAM                         &16$\times $4GB; 2.67MT$/$s\\
\hline
GPU (number and type)                         & $8 \times$ NVIDIA V100 32G\\
\hline
CUDA version                  & 11.3\\                          \hline
Programming language                 & Python 3.9\\ 
\hline
Deep learning framework & Pytorch (Torch 1.12, torchvision 0.13.0) \\

\hline
\end{tabular}
\end{table}

\textbf{Data augmentation} 8 data augmentation techniques are used sequentially, which are RandSpatialCrop, RandAxisFlip, and RandRotate90 with probability 0.5, RandGaussianNoise with probability 0.5, RandAdjustContrast, RandGaussianSmooth, and RandHistogramShift with probability 0.25, RandZoom with $min\_zoom$ as 0.5 and $max\_zoom$ as 2.

\textbf{Hyper-parameters}
In the unsupervised classification step of the automatic cell classification pipeline, the threshold of Saturation $\theta$ is set to 0.1 and the range of Value ($\alpha_s$,$\alpha_l$) is set to (0.1,0.6). The area threshold to divide large/small cells is denoted as $\sigma$ and is set to 8000. In the ConvNeXt-Stardist network, the number of rays $R$ is set to 32. In the non-maximum suppression of the ConvNeXt-Stardist, we set the object probability threshold to 0.5 and the IOU threshold to 0.4. In the marker-controlled watershed algorithm of ConvNeXt-Hover, we set the probability threshold of cell pixels to 0.6.

\begin{table*}[!htbp]
\centering
\small
\caption{Training protocols.}\label{table:training}
\begin{tabular}{c|c|c} 
~ & Cell image classification model & Cell segmentation models \\
\hline
 Network initialization  &  ImageNet-pretrained ResNet18 & ImageNet-pretrained ConvNeXt-small \\
\hline
 Batch size                    & 64  & 16 \\
\hline
 Patch size & $224\times224$  & $512\times 512$\\ 
\hline
Total epochs & 20 & 800 \\
\hline
Optimizer          & SGD    & AdamW     \\ 
\hline
 Initial learning rate  & 0.001 & 0.0001\\
\hline
 Lr decay schedule & decay by a factor of 10 every 7 epochs & halved by 200 epochs\\
\hline
 Training time                                           & 17 minutes & 72 hours \\ 
\hline
 Stopping criteria  &  Train for 20 epochs & Train for 800 epochs  \\ 
\hline
Model selection criteria   &  Accuracy on the validation set &  F1 score on the validation set\\ \hline
\end{tabular}
\end{table*}

\textbf{Pre-processing}
During the training, we randomly sample $512\times512$ image patches. In the inference stage, we use a sliding window to crop image patches of size $512\times512$ with a step of $384$. Data augmentations are not used in the testing.

\textbf{Post-processing}
For the ConvNeXt-stardist output, we perform non-maximum suppression (NMS)~\cite{ren2015faster} to obtain the cells with the highest object probabilities and discard highly-overlapping objects. For the ConvNeXt-Hover, we perform the marker-controlled watershed algorithm~\cite{cheng2008segmentation} using the predicted CP map and HV maps to separate overlapped cells.

\section{Results and discussion}
\subsection{Performance of the deep classification model}
\begin{figure*}[h]
\centering
\includegraphics[width=1.0\linewidth]{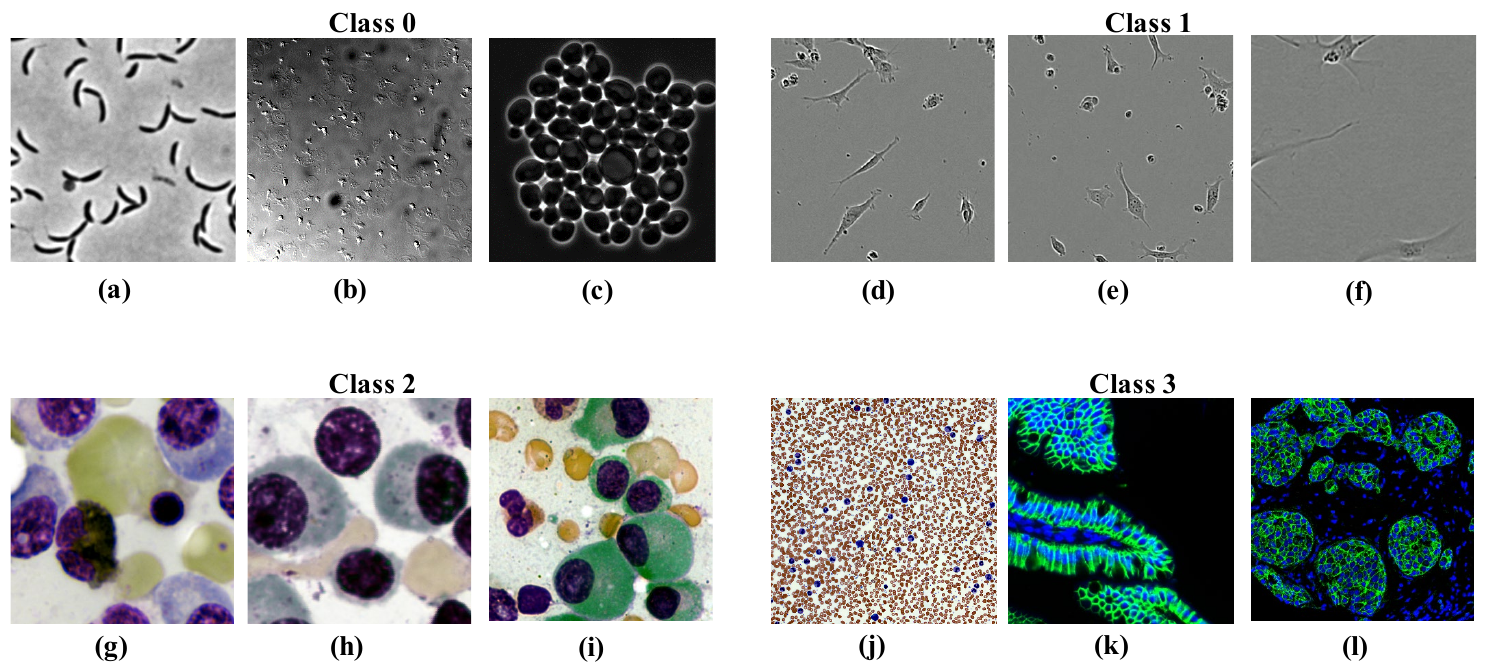}
\caption{Selected results of cell image classification on the validation set.}
\label{cls_result}
\end{figure*}

The classification model used in this method is ResNet18~\cite{he2016deep}, and the classification accuracies for four classes are 99.48\%, 84.51\%, 100\% and 98.89\%, respectively. The overall classification accuracy on the validation set is 97.91\%. Note that the overall classification accuracy is not the mean value of the accuracy of the four classes. It is defined as the percentage of correctly classified image numbers over all the validation images. Figure~\ref{cls_result} shows some classification results using the well trained deep classification model. 
\subsection{Effectiveness of class-wise finetuning and model combination}
We conduct ablation studies on the validation set and the tuning set to evaluate the effectiveness of the class-wise multi-stream framework. In Table~\ref{Table.4}, $F^1_{0}$, $F^1_{1}$, $F^1_{2}$, $F^1_{3}$ are the F1 score for four classes on validation set. $F^1_{v}$ and $F^1_{t}$ are the mean F1 score on the validation set and tuning Set. `ConvNeXt-Stardist-Pretrained' and `ConvNeXt-Hover-Pretrained' represent the  ConvNeXt-Stardist and ConvNeXt-Hover models trained on all the labeled images. `ConvNeXt-Stardist-Finetuned' and `ConvNeXt-Hover-Finetuned' are the models of two segmentation methods fine-tuned on the images of different classes. Firstly, comparing the mean F1 score on the tuning set of two pre-trained models and the fine-tuned models, the class-wise fine-tuning can improve the segmentation performance by 1.96\% or 5.81\%, respectively. The results show that the class-wise fine-tuning is useful in this multi-modal cell segmentation task. Secondly, compared with the tuning set F1 score of the segmentation models `ConvNeXt-Stardist-Finetuned' and `ConvNeXt-Hover-Finetuned',  our framework achieves 1.55\% and 1.91\% F1 score improvement, respectively. The improvement suggests that combining different segmentation models that are trained for different classes is effective. On the validation set, `ConvNeXt-Stardist' shows better segmentation results on Class 0,3, and `ConvNeXt-Hover' performs better on Class 1,2. However, since we use the F1-scores on the tuning set $F^1_t$ as the model selection metrics. In our final solution, we use ConvNeXt-Stardist for Class 0,2,3 and ConvNeXt-Hover for Class 1. `Ours-(Hover for Class 2)' denotes changing the Class 2 model in `Ours' to ConvNeXt-Hover, which is slightly worse than `Ours' in $F^1_t$. 

\begin{table*}[!htb]
\centering
\small
\caption{Effectiveness of class-wise multi-stream cell segmentation framework. 'Pretrained' mean that the model is trained using all the labeled images, including the competition dataset and public datasets. 'Finetuned' represents that the models are fine-tuned using images of different classes. 'Ours' is the final model that uses ConvNeXt-Stardist-Finetuned for Class 0,2,3 and ConvNeXt-Hover-Finetuned for Class1. }
\begin{tabular}{c|c|c|c|c|c|c}
     & $F^1_{0}$ & $F^1_{1}$& $F^1_{2}$ & $F^1_{3}$ & $F^1_{v}$ & $F^1_{t}$ \\ \hline
ConvNeXt-Stardist-Pretrained & 0.9105& 0.7626 & 0.8177 & 0.8582 & 0.8771 & 0.8444                    \\
ConvNeXt-Hover-Pretrained    & 0.8930& 0.8572 & 0.8340 & 0.8180 & 0.8552 & 0.8023 \\ \hline
ConvNeXt-Stardist-Finetuned  & 0.9320 & 0.8281 & 0.8625 & 0.8647& 0.8866 & 0.8640                    \\
ConvNeXt-Hover-Finetuned  & 0.9206  & 0.8565                    & 0.8927  & 0.8454                    & 0.8869                    & 0.8604                    \\ \hline
Ours-(Hover for Class 2) & 0.9320 & 0.8565 & 0.8927 & 0.8647 & 0.8897 & 0.8764 \\
Ours with unlabeled data & 0.8821 & 0.8534 & 0.8777 & 0.8709 & 0.8735 & 0.8695 \\
Ours  & 0.9320& 0.8565                    & 0.8625                    & 0.8647                    & 0.8875                    & \textbf{0.8795}                   
\end{tabular}
\label{Table.4}
\end{table*}

\subsection{Quantitative results on the tuning set}
The F1 score of our method on the tuning set is 0.8795. Note that the segmentation models in our pipeline are trained with only the labeled data, without using any unlabeled data. For training the deep cell image classifier, we use unlabeled data. We find that using or not using unlabeled data will achieve exactly the same classification results on the tuning set. When the segmentation models are fixed, the same image classification results lead to the same cell segmentation scores. We hope that the deep image classifier can enjoy better generalization ability on testing sets, so we still choose to use unlabeled data for the classifier training. \\
\textbf{Use of unlabeled data in cell segmentation} We perform the pseudo-label~\cite{lee2013pseudo} method to use the unlabeled data. The pseudo labels are obtained by inferring unlabeled images with the ConvNeXt-Finetuned models. Then these unlabeled images are classified into four classes by our deep classification model and mixed with the labeled images. These mixed images and their pseudo-labels are used for finetuning the ConvNeXt-Stardist-Finetuned/ConvNeXt-Hover-Finetuned models. The segmentation results are shown as `Ours with unlabeled data' in Table~\ref{Table.4}. The results show that using the unlabeled data with the Pseudo-labeling method does not enhance the segmentation performance in our solution.

\subsection{Qualitative results on the validation set}
\begin{figure*}[h]
\centering
\includegraphics[width=0.9\linewidth]{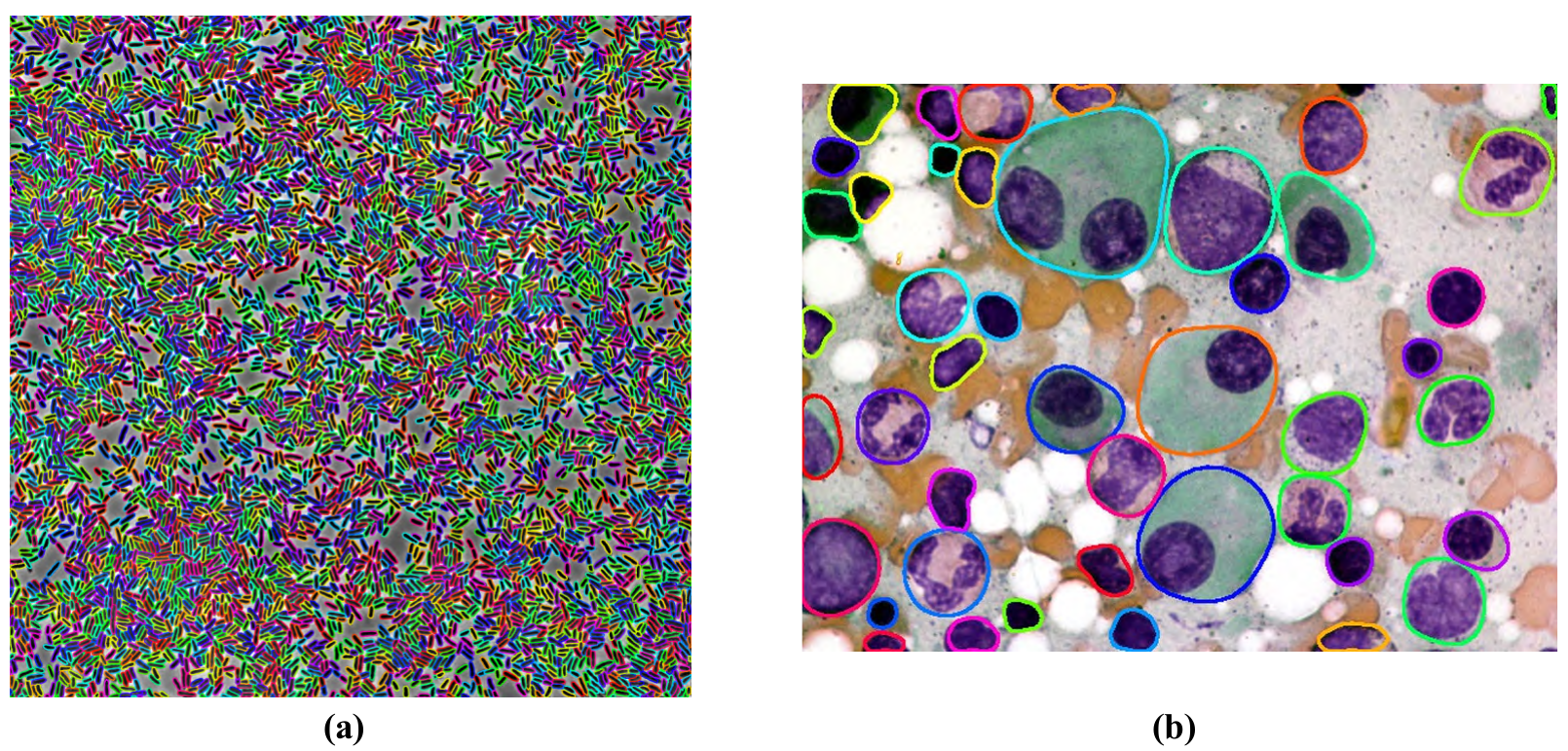}
\caption{Good segmentation results.}
\label{good_seg}
\end{figure*}

\begin{figure*}[h]
\centering
\includegraphics[width=0.9\linewidth]{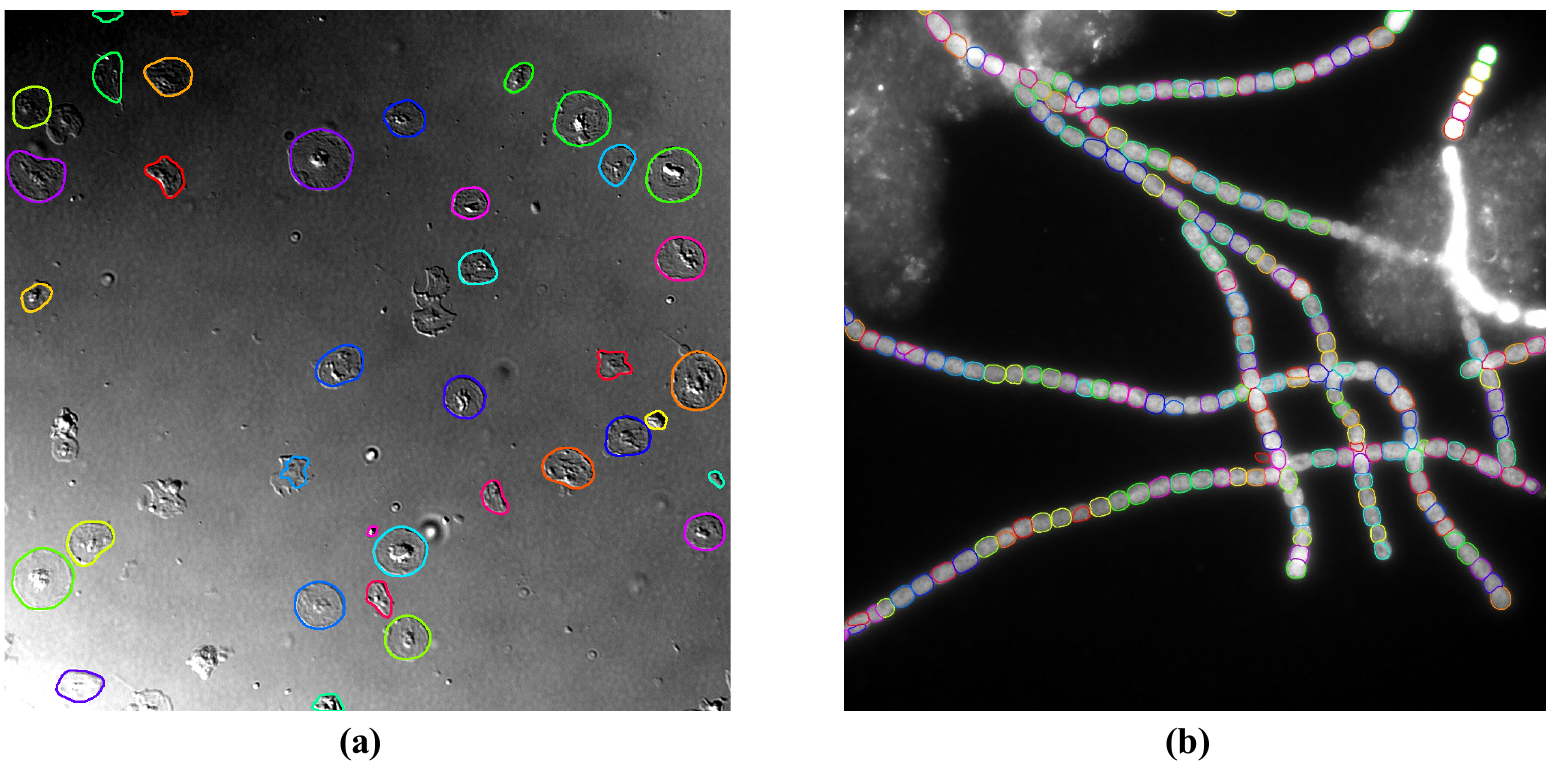}
\caption{Poor segmentation results.}
\label{bad_seg}
\end{figure*}
The visual results in Figure~\ref{good_seg} show that our method works well not only on the microscopy images with densely distributed cells, but also on the images with sparsely distributed objects. However, in some cell types like Figure~\ref{bad_seg}(a), our method may miss some cells. Meanwhile, our method may predict false positives or fail to segment some cells for the images with bright foregrounds and dark backgrounds, like Figure~\ref{bad_seg}(b). The failures may be due to the lack of training data that belong to this type.

\subsection{Segmentation efficiency on the tuning set}
The out-of-tolerance running time of our method on all the 101 cases in the tuning set is zero in our local workstation. Table~\ref{Table.5} shows the time tolerance and real running time of some images in the tuning set. With the increase of the image size, the real running time is increasing but within the time tolerance.

\begin{table*}[!htb]
\centering
\small
\caption{The running time of selected images in the tuning set. `Out-of-Tol. RT' denotes the running time out of the time tolerance. }
\begin{tabular}{c|c|c|c|c}
Name        & Size       & Time tolerance (s) & Real Running Time (s) & Out-of-Tol. RT (s) \\ \hline
cell\_00001 & 640*480    & 10                 & 9.36                 & 0                    \\\hline
cell\_00009 & 3000*3000  & 90                 & 13.52                & 0                    \\\hline
cell\_00033 & 1266*944   & 12                 & 9.72                 & 0                    \\\hline
cell\_00073 & 2048*2048  & 42                 & 11.9                 & 0                    \\\hline
cell\_00101 & 10496*8415 & 883                & 388              & 0                    \\ 
\end{tabular}
\label{Table.5}
\end{table*}

\subsection{Results on the final testing set}

\begin{table}[!htb]
\centering
\small
\caption{The median and mean F1-score for each modality of our solution on the final testing set. `BF', `DIC', `Fluo', `PC' represents Brightfield, Differential interference contrast, Fluorescent and Phase-contrast modalities. }
\begin{tabular}{c|c|c|c|c|c}
Team   Name               & Median   F1-All & Median   F1-BF & Median   F1-DIC & Median   F1-Fluo & Median   F1-PC \\ \hline
\multirow{3}{*}{sribdmed} & 0.8474          & 0.91           & 0.7496          & 0.6645           & 0.8878         \\ \cline{2-6}
& Mean F1-All     & Mean F1-BF     & Mean F1-DIC     & Mean F1-Fluo     & Mean F1-PC     \\ \cline{2-6}
& 0.7846          & 0.9002         & 0.6903          & 0.6963           & 0.8044    \\   \hline 
\end{tabular}
\end{table}
In Table 6, our solution shows outstanding segmentation performance on Brightfield and Phase-contrast images. The mean F1-score and median F1-score are all over 0.8. However, our solution has difficulty in segmenting the cells from Differential interference contrast and Fluorescent images precisely. Overall, our solution obtains competitive results for processing multi-modality images with the Median F1-All score of 0.8474 and the Mean F1-ALL score of 0.7846.

\subsection{Limitation and future work}
In the future, we may develop more semi-supervised methods to further enhance the segmentation performance of our proposed method. Besides, it is possible to improve the segmentation by using multiple decoders or other decoder structures. We may investigate if using more low-level attributes to define more categories can further improve the segmentation performance.

\section{Conclusion}
In this work, we introduce an automatic cell classification pipeline and a class-wise multi-stream cell segmentation framework to solve the multi-modal cell segmentation task. The proposed method shows great generalization and performance on different types of cells and maintains high efficiency. We believe that our framework can serve as a strong baseline model for the cell segmentation task.

\section*{Acknowledgement}
The authors of this paper declare that the segmentation method they implemented for participation in the NeurIPS 2022 Cell Segmentation challenge has not used any private datasets other than those provided by the organizers and the official external datasets and pretrained models. The proposed solution is fully automatic without any manual intervention. This work is supported by Chinese Key-Area Research and Development Program of Guangdong Province (2020B0101350001), the National Natural Science Foundation of China (No.62102267), Guangdong Basic and Applied Basic Research Foundation (2023A1515011464), and the Guangdong Provincial Key Laboratory of Big Data Computing, The Chinese University of Hong Kong, Shenzhen. 

\bibliographystyle{unsrt}
\bibliography{ref}

\begin{thebibliography}{10}

\bibitem{prangemeier2020attention}
Tim Prangemeier, Christoph Reich, and Heinz Koeppl.
\newblock Attention-based transformers for instance segmentation of cells in
  microstructures.
\newblock In {\em International Conference on Bioinformatics and Biomedicine
  (BIBM)}, pages 700--707. IEEE, 2020.

\bibitem{lee2022cellseg}
Michael~Y Lee, Jacob~S Bedia, Salil~S Bhate, Graham~L Barlow, Darci Phillips,
  Wendy~J Fantl, Garry~P Nolan, and Christian~M Sch{\"u}rch.
\newblock Cellseg: a robust, pre-trained nucleus segmentation and pixel
  quantification software for highly multiplexed fluorescence images.
\newblock {\em BMC Bioinformatics}, 23(1):1--17, 2022.

\bibitem{sun2020deep}
Jing Sun, Attila T{\'a}rnok, and Xuantao Su.
\newblock Deep learning-based single-cell optical image studies.
\newblock {\em Cytometry Part A}, 97(3):226--240, 2020.

\bibitem{prangemeier2020multiclass}
Tim Prangemeier, Christian Wildner, Andr{\'e}~O Fran{\c{c}}ani, Christoph
  Reich, and Heinz Koeppl.
\newblock Multiclass yeast segmentation in microstructured environments with
  deep learning.
\newblock In {\em Conference on Computational Intelligence in Bioinformatics
  and Computational Biology (CIBCB)}, pages 1--8. IEEE, 2020.

\bibitem{leygeber2019analyzing}
Markus Leygeber, Dorina Lindemann, Christian~Carsten Sachs, Eugen Kaganovitch,
  Wolfgang Wiechert, Katharina N{\"o}h, and Dietrich Kohlheyer.
\newblock Analyzing microbial population heterogeneity—expanding the toolbox
  of microfluidic single-cell cultivations.
\newblock {\em Journal of Molecular Biology}, 431(23):4569--4588, 2019.

\bibitem{lou2022pixel}
Wei Lou, Haofeng Li, Guanbin Li, Xiaoguang Han, and Xiang Wan.
\newblock Which pixel to annotate: a label-efficient nuclei segmentation
  framework.
\newblock {\em IEEE Transactions on Medical Imaging}, 2022.

\bibitem{zhou2021ssmd}
Hong-Yu Zhou, Chengdi Wang, Haofeng Li, Gang Wang, Shu Zhang, Weimin Li, and
  Yizhou Yu.
\newblock Ssmd: semi-supervised medical image detection with adaptive
  consistency and heterogeneous perturbation.
\newblock {\em Medical Image Analysis}, 72:102117, 2021.

\bibitem{schmidt2018cell}
Uwe Schmidt, Martin Weigert, Coleman Broaddus, and Gene Myers.
\newblock Cell detection with star-convex polygons.
\newblock In {\em International Conference on Medical Image Computing and
  Computer-Assisted Intervention}, pages 265--273. Springer, 2018.

\bibitem{graham2019hover}
Simon Graham, Quoc~Dang Vu, Shan E~Ahmed Raza, Ayesha Azam, Yee~Wah Tsang,
  Jin~Tae Kwak, and Nasir Rajpoot.
\newblock Hover-net: Simultaneous segmentation and classification of nuclei in
  multi-tissue histology images.
\newblock {\em Medical Image Analysis}, 58:101563, 2019.

\bibitem{liu2022convnet}
Zhuang Liu, Hanzi Mao, Chao-Yuan Wu, Christoph Feichtenhofer, Trevor Darrell,
  and Saining Xie.
\newblock A convnet for the 2020s.
\newblock In {\em Proceedings of the IEEE/CVF Conference on Computer Vision and
  Pattern Recognition}, pages 11976--11986, 2022.

\bibitem{stringer2021cellpose}
Carsen Stringer, Tim Wang, Michalis Michaelos, and Marius Pachitariu.
\newblock Cellpose: a generalist algorithm for cellular segmentation.
\newblock {\em Nature Methods}, 18(1):100--106, 2021.

\bibitem{cutler2022omnipose}
Kevin~J Cutler, Carsen Stringer, Teresa~W Lo, Luca Rappez, Nicholas Stroustrup,
  S~Brook~Peterson, Paul~A Wiggins, and Joseph~D Mougous.
\newblock Omnipose: a high-precision morphology-independent solution for
  bacterial cell segmentation.
\newblock {\em Nature Methods}, pages 1--11, 2022.

\bibitem{Sartorius}
Sartorius.
\newblock Sartorius - cell instance segmentation.
\newblock [Online].
\newblock
  \url{https://www.kaggle.com/competitions/sartorius-cell-instance-segmentation}.

\bibitem{lee2013pseudo}
Dong-Hyun Lee et~al.
\newblock Pseudo-label: The simple and efficient semi-supervised learning
  method for deep neural networks.
\newblock In {\em Workshop on Challenges in Representation Learning, ICML},
  volume~3, page 896, 2013.

\bibitem{he2016deep}
Kaiming He, Xiangyu Zhang, Shaoqing Ren, and Jian Sun.
\newblock Deep residual learning for image recognition.
\newblock In {\em Proceedings of the IEEE Conference on Computer Vision and
  Pattern Recognition}, pages 770--778, 2016.

\bibitem{ronneberger2015u}
Olaf Ronneberger, Philipp Fischer, and Thomas Brox.
\newblock U-net: Convolutional networks for biomedical image segmentation.
\newblock In {\em International Conference on Medical Image Computing and
  Computer-assisted Intervention}, pages 234--241. Springer, 2015.

\bibitem{ren2015faster}
Shaoqing Ren, Kaiming He, Ross Girshick, and Jian Sun.
\newblock Faster r-cnn: Towards real-time object detection with region proposal
  networks.
\newblock {\em Advances in Neural Information Processing Systems}, 28, 2015.

\bibitem{cheng2008segmentation}
Jierong Cheng, Jagath~C Rajapakse, et~al.
\newblock Segmentation of clustered nuclei with shape markers and marking
  function.
\newblock {\em IEEE Transactions on Biomedical Engineering}, 56(3):741--748,
  2008.

\bibitem{javer2018open}
Avelino Javer, Michael Currie, Chee~Wai Lee, Jim Hokanson, Kezhi Li,
  C{\'e}line~N Martineau, Eviatar Yemini, Laura~J Grundy, Chris Li, QueeLim
  Ch’ng, et~al.
\newblock An open-source platform for analyzing and sharing worm-behavior data.
\newblock {\em Nature Methods}, 15(9):645--646, 2018.

\bibitem{ljosa2012annotated}
Vebjorn Ljosa, Katherine~L Sokolnicki, and Anne~E Carpenter.
\newblock Annotated high-throughput microscopy image sets for validation.
\newblock {\em Nature Methods}, 9(7):637--637, 2012.

\bibitem{deng2009imagenet}
Jia Deng, Wei Dong, Richard Socher, Li-Jia Li, Kai Li, and Li~Fei-Fei.
\newblock Imagenet: A large-scale hierarchical image database.
\newblock In {\em IEEE Conference on Computer Vision and Pattern Recognition},
  pages 248--255, 2009.

\end{thebibliography}

\end{document}